\documentclass[twoside,leqno,twocolumn]{article}
\usepackage{ltexpprt}
\usepackage{graphicx}
\usepackage{algorithmic}
\usepackage[cmex10]{amsmath}
\usepackage{amssymb}

\usepackage{enumitem}
\usepackage{times} %
\usepackage{helvet} %
\usepackage{courier} %
\usepackage{multirow} %
\usepackage{epsfig} %
\usepackage{algorithm} %
\usepackage{verbatim} %

\newcommand{\Keywords}[1]{\par\noindent{\small{\bf Keywords\/}: #1}}
\def\etal{{\em et al.\/}\,}
\newcommand{\Us}{\mathcal{U}}
\newcommand{\Is}{\mathcal{V}}
\newcommand{\Loss}{\mathcal{L}}

\newcommand{\W}{\mathbf{W}}

\newcommand{\X}{\mathbf{X}}

\newcommand{\R}{\mathbf{R}}

\def\1{{\mathbf 1}}
\def\0{{\mathbf 0}}

\begin{document}

\title{Selective Transfer Learning for Cross Domain Recommendation}

\newcommand*\samethanks[1][\value{footnote}]{\footnotemark[#1]}
\author{Zhongqi Lu\thanks{Hong Kong University of Science \& Technology. \{zluab, ezhong, skyezhao, wxiang, weikep, qyang\}@cse.ust.hk}
\and
Erheng Zhong\samethanks
\and
Lili Zhao\samethanks
\and
Evan Xiang\samethanks
\and
Weike Pan\samethanks
\and
Qiang Yang\samethanks
}
\date{}

\maketitle

\begin{abstract} \small\baselineskip=9pt
{\bf C}ollaborative {\bf f}iltering (CF) aims to predict users' ratings on items according to historical user-item preference data. In many real-world applications, preference data are usually sparse, which would make models overfit and fail to give accurate predictions.
Recently, several research works show that by transferring knowledge from some manually selected source domains, the data sparseness problem could be mitigated.
However for most cases, parts of source domain data are {\em not consistent} with the observations in the target domain, which may misguide the target domain model building.
In this paper, we propose a novel criterion based on empirical prediction error and its variance to better capture the consistency across domains in CF settings. Consequently, we embed this criterion into a boosting framework to perform {\em selective} knowledge transfer.
Comparing to several state-of-the-art methods, we show that our proposed selective transfer learning framework can significantly improve the accuracy of rating prediction tasks on several real-world recommendation tasks.
\end{abstract}

\vspace{0.5cm}

\section{Introduction}\label{sec:intro}

Recommendation systems attempt to recommend items (e.g., movies, TV, books, news, images, web pages, etc.) that are likely to be of interest to users. As a state-of-the-art technique for recommendation systems, collaborative filtering aims at predicting users' ratings on a set of items based on a collection of historical user-item preference records.
In the real-world recommendation systems, although the item space is often very large, users usually rate only a small number of items. Thus, the available rating data can be extremely sparse for each user, which is especially true for new online services.
Such data sparsity problem may make CF models overfit the limited observations and result in low-quality predictions.

In recent years, different transfer learning techniques have been developed to improve the performance of learning a model via reusing some information from other relevant systems for collaborative filtering~\cite{/ijcai/libin09,/uai/ZhangCY10}.
And with the increasing understandings of auxiliary data sources, some works (like ~\cite{DBLP:conf/aaai/EldardiryN11,DBLP:conf/sdm/ShiPGY12}) start to explore data from multiple source domains to achieve more comprehensive knowledge transfer.
However, these previous methods over-trust the source data and assume that the source domains follow the very similar distributions with the target domain, which is usually not true in the real-world applications, especially under the cross domain CF settings.
For example, in a local music rating web site, natives may give trustful ratings for the traditional music; while in an international music rating web site, the ratings on those traditional music could be diverse due to the culture differences: those users with good culture background would constantly give trustful ratings, others could be inaccurate. If the target domain task is the music recommendation of a startup local web site, obviously we do not want all the International web site's data as source domain without selection. To better tackle the cross domain CF problems, we face the challenge to tell how consistent the data of target and source domains are and adopt only those consistent source domain data while transferring knowledge.

Several research works (like ~\cite{DBLP:conf/icml/DaiYXY07}) have been proposed to perform instance selection across domains for classification tasks based on empirical error. But they cannot be adopted to solve CF problems directly. Especially when the target domain is sparse, because of the limited observations of user's ratings on the items in the target domain, getting a low empirical error occasionally in the target domain does not mean the source domains are truly helpful in building a good model. In other words, the inconsistent knowledge from source domains may dominate the target domain model building and happen to fit the few observations in the target domain, which gives high accuracy unacceptably.

We take careful analysis on this problem and in our observation on the music rating example, some users, such as domain experts, follow standard criteria to rate and hence share a consistent distribution over the mutual item set across domains. And further, we find this consistency can be better described by adding the variance of empirical error produced by the model. The smaller the variance of empirical error on predictions for a user, the more likely this user is consistent with those from other domains. And we would like to adopt those who are more likely to share consistent preferences to perform knowledge transfer across domains.
Based on this observation, we propose a new criterion using both empirical error and its variance to capture the consistency between source and target domains. As an implementation, we embed this criterion into a boosting framework and propose a novel selective transfer learning approach for collaborative filtering (STLCF).
STLCF works in an iterative way to adjust the importance of source instances, where those source data with low empirical error as well as low variance will be selected to help build target models.

Our main contributions are summarized as follows:
\begin{itemize}[noitemsep,topsep=0pt,parsep=0pt,partopsep=0pt]
\item First, we find that selecting consistent auxiliary data for the target domain is important for the cross-domain collaborative filtering, while the consistency between source and target domain is influenced by multiple factors. To describe these factors, we propose a novel  criterion, based on both empirical error and its variance.
\item Second, we propose a {\em selective} transfer learning framework for collaborative filtering - an extension of the boosting based transfer learning algorithm that take the above factors into consideration, so that the sparseness issue in the CF problems can be better tackled.
\item Third, the proposed framework is general, where different base models can be embedded. We propose an implementation based on Gaussian probability latent semantic analysis, which demonstrates the proposed framework can solve the sparseness problem on various real-world applications.
\end{itemize}


\section{Preliminaries}
\subsection{Problem Settings}

\begin{figure}
\includegraphics[width=3.2in]{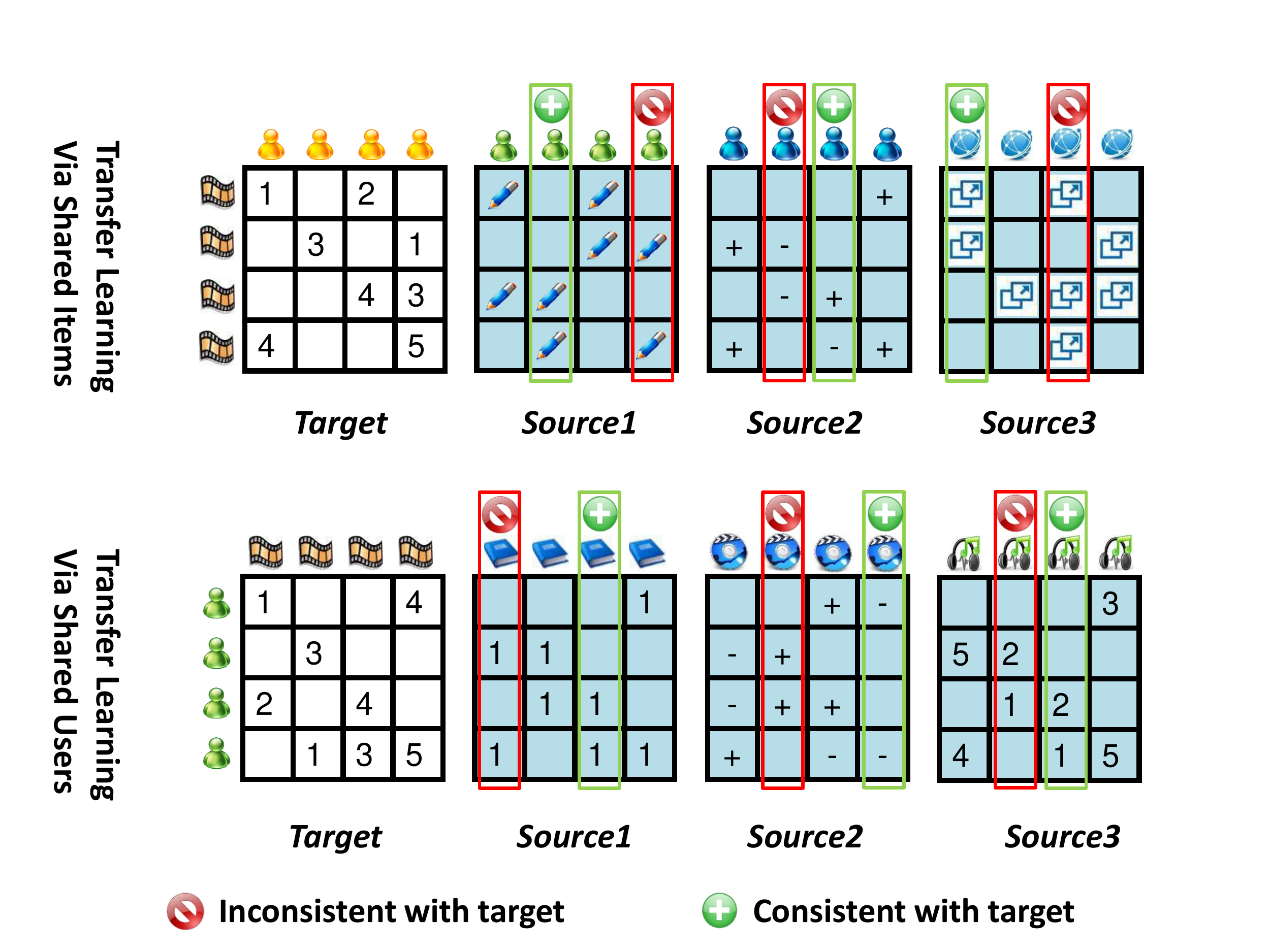}
\caption{\footnotesize Selective transfer learning with multiple sources. The first row illustrates the case where items that are in the target domain also appear in the source domains. A real-world example is the rating prediction for the movies that appear in several web sites in various forms; the second row illustrates the case where users that are in the target domain also appear in the source domains. A real-world example is the Douban recommendation system, which provides music, book and movie recommendation for users.}\label{fig:illustration}
\vspace{-5mm}
\end{figure}

Suppose that we have a target task $\mathcal{D}$ where we wish to solve the rating prediction problem. Taking the regular recommendation system for illustration, $\mathcal{D}$ is associated with $m_{d}$ users and $n_{d}$ items denoted by $\Us^{d}$ and $\Is^{d}$ respectively. In this task, we observe a sparse matrix $\X^{(d)}\in \mathbb{R}^{m_d\times n_{d}}$ with entries $x^d_{ui}$. Let $\R^{(d)}=\{(u,i,r):r=x^d_{ui},\mbox{where } x^d_{ui}\ne 0\}$ denote the set of observed links in the system. For the rating recommendation system, $r$ can either take numerical values, for example $[1,5]$, or binary values $\{0, 1\}$. We aim to transfer knowledge from other $N$ source domains $\mathcal{S}=\{S^t\}_{t=1}^{N}$ with each source domain $S^t$ contains $m^t_s$ users and $n^t_s$ items denoted by $\Us^{s^t}$ and $\Is^{s^t}$. Similar to the target domain, each source domain $S^t$ contain sparse matrices $\X^{(s^t)}\in \mathbb{R}^{m_{s^t}\times n_{s^t}}$ and observed links $\R^{(s^t)}=\{(u,i,r):r=x^{s^t}_{ui},\mbox{ where } x^{s^t}_{ui}\ne 0\}$.

The settings of STLCF are illustrated in Figure \ref{fig:illustration}. We adopt a setting commonly used in transfer learning for collaborative filtering: either the items or the users that are in the target domain also appear in the source domains. In the following derivation and description of our STLCF model, for the convenience of interpretation, we focus on the case that the user set is shared by both target domain and the source domains. The case that the item set is shared can be easily tackled in a similar manner.

Under the assumption that the observation $\R^{(\{d, s^t\})}$ is obtained with $u$ and $i$ being independent, we formally define a co-occurrence model in both the target and the source domains to solve the collaborative filtering problem:
\begin{eqnarray}
\nonumber
Pr(x^{\{d, s^t\}}_{ui}=r,u,i) \!\!\!\!\!\!&=&\!\!\!\!\!\! Pr(u) Pr(i \mid u) Pr(x^{\{d, s^t\}}_{ui}=r \mid u,i) \\
\nonumber
\!\!\!\!\!\!&=&\!\!\!\!\!\! Pr(u) Pr(i) Pr(x^{\{d, s^t\}}_{ui}=r \mid u,i) \\
\nonumber
\!\!\!\!\!\!&\propto&\!\!\!\!\!\! Pr(x^{\{d, s^t\}}_{ui}=r \mid u,i)
\end{eqnarray}

In the following, based on Gaussian probabilistic latent semantic analysis (GPLSA), we first briefly present a transfer learning model for collaborative filtering - transferred Gaussian probabilistic latent semantic analysis (TGPLSA) as an example, which is designed to integrate into our later proposed framework as a base model. After that, we present our selective transfer learning for collaborative filtering (STLCF) to perform knowledge transfer by analyzing the inconsistency between the observed data in target domain and the source domains. Careful readers shall notice that other than the TGPLSA example, STLCF is compatible to use various generative models as the base model.

\subsection{Collaborative Filtering via Gaussian Probabilistic Latent Semantic Analysis (GPLSA)}
For every user-item pair, we introduce hidden variables $Z$ with latent topics $z$, so that user $u$ and item $i$ are rendered conditionally independent. With observations of item set $\Is$, user set $\Us$ and rating set $\R$ in the source domain, we define a model as:
\begin{eqnarray}
\nonumber
Pr(x_{ui} = r | u, i) = \sum_{z} Pr(x_{ui} = r \mid i, z) Pr(z \mid u)
\end{eqnarray}

We further investigate the use of a Gaussian model for estimating $p(x_{ui} = r | u, i)$ by introducing $\mu_{iz} \in \mathcal{R}$ for the mean and $\sigma_{iz}^2$ for the variance of the ratings. With these, we define a Gaussian mixture model for a single domain as:
\begin{eqnarray}
\nonumber
Pr(x_{ui} = r | u, i) = \sum_{z} Pr(z | u) Pr(r; \mu_{iz}, \sigma_{iz})
\end{eqnarray}
where $Pr(z |u)$ is the topic distribution over users, and $Pr(r; \mu_{iz}, \sigma_{iz})$ follows a Gaussian distribution.

Maximum likelihood estimation amounts to minimize:
\begin{eqnarray}\label{eq:likelihood}
\Loss = -  \sum_{r \in R} \sum_{x_{ui} \in \X} \log [ Pr(x_{ui} = r \mid u,i ; \theta) ]
\end{eqnarray}
where $\theta$ is a generic parameter referring to a particular model.

Next, we extend GPLSA to the cross domain context to achieve transfer learning for collaborative filtering (TLCF).

\subsection{Transfer Learning for Collaborative Filtering (TLCF)}
When the target data $\X^{(d)}$ is sparse, GPLSA may overfit the limited observed data.
Following the similar idea in \cite{DBLP:conf/sigir/XueDYY08}, we extend GPLSA to the Transferred Gaussian Probabilistic Latent Semantic Analysis (TGPLSA) model. Again we use $s$ to denote index of the source domain where the knowledge come from, and $d$ to denote the index of the target domain where the knowledge is received.
For simplicity, we present the work with one source domain, and this model can be easily extended to multiple source domains. Moreover, we assume all the users appear in both the source domains and the target domain. Such scenarios are common in the real-world systems, like Douban\footnote{http://www.douban.com - a widely used online service in China, which provides music, book and movie recommendations.}.


Similar to the approach in \cite{DBLP:conf/sigir/XueDYY08}, the TGPLSA jointly learn the two models for both the source domain and the target domain using a relative weight parameter\footnote{$\lambda \in (0,1)$, which is introduced to represent the tradeoff of source and target information.} $\lambda$. Since the item sets $\Is^s$ and $\Is^d$ are different or even disjoint with each other, there could be inconsistency across domains as we discussed in Section \ref{sec:intro}. Clearly, the more consistent source and target domains are, the more help target task could get from source domain(s). We are motivated to further analyze this inconsistency in our work by learning item weight vectors $\W^{s} = \{w_i^s\}_{i=1}^{n^s}$ and $\W^{d} = \{w_i^d\}_{i=1}^{n^d}$ of the instances in source and target domain respectively. Then, the objective function in Eq.(\ref{eq:likelihood}) can be extended as:
\begin{eqnarray}\label{eq:tplsa_likelihood}
\\
\nonumber
\Loss = - \sum_{r \in R} ( \lambda \sum_{x_{ui}^s \in \X^s} \log ( w_i^s \cdot Pr(x_{ui}^s = r \mid u^s,i^s ; \theta^s) )  \\\nonumber
+ (1-\lambda) \sum_{x_{ui}^d \in \X^d} \log ( w_i^d  \cdot Pr(x_{ui}^d = r \mid u^d,i^d ; \theta^{d}) ) )
\end{eqnarray}

We adopt the expectation-maximization (EM) algorithm, a standard method for statistical inference, to find the maximum log-likelihood estimates of Eq.(\ref{eq:tplsa_likelihood}). Details of derivations can be found in the appendix.


\begin{algorithm}[tb]
\caption{\footnotesize Selective TLCF.}
\begin{algorithmic}

\STATE {\bfseries Input:} $\X^{d}$, $\X^{s}$, $T$\\
$\X^{d} \in \mathbb{R}^{m\times n^d}$: the target training data \\
$\X^{s} \in \mathbb{R}^{m\times n^s}$: the auxiliary source data\\
$G$: the weighted TLCF model wTGPLSA\\
$T$: number of boosting iterations \\

\STATE

\STATE {\bfseries Initialize:} Initialize $\W^s : w_i^s\leftarrow\frac{1}{n^s}$, $\W^d : w_i^d\leftarrow\frac{1}{n^d}$

\FOR{ $iter$ = 1 to $T$}

\STATE {\bfseries Step 1:} Apply $G$ to generate a weak learner $G(\X^{d}, \X^{s}, \W^{d}, \W^{s})$ that minimize Eq.(\ref{eq:tplsa_likelihood}) 

\STATE {\bfseries Step 2:} Get weak hypothesis for both the $d$ and $s$ domains $h^{iter} : \X^{d}, \X^{s}\rightarrow \hat \X^{d}, \hat \X^{s}$

\STATE {\bfseries Step 3:} Calculate empirical error $E^{d}$ and $E^{s}$ using Eq.(\ref{eq:vpb_loss})

\STATE {\bfseries Step 4:} Calculate fitness weight $\beta^{s_k}$ for each source domain $s_k$ using Eq.(\ref{eq:beta})

\STATE {\bfseries Step 5:} Choose model weight $\alpha^{iter}$ via Eq.(\ref{eq:alpha})

\STATE {\bfseries Step 6:} Update source item weight $\W^{s}$ via Eq.(\ref{eq:wi_s})

\STATE {\bfseries Step 7:} Update target item weight $\W^{d}$  via Eq.(\ref{eq:wi})

\ENDFOR

\STATE {\bfseries Output:} Hypothesis $Z = H(\X^{(d)}) = \Sigma_{t=1}^T \alpha^t h^t(\X^{(d)})$

\end{algorithmic}
\label{algorithm:VPB-TLCF}
\end{algorithm}

\section{Selective TLCF} \label{sec:STLCF}

As we have discussed before, using the source domain data without selection may harm the target domain learning.
By proposing the selective knowledge transfer with the novel factors (empirical error and variance of empirical error), we come up with the details of Selective Transfer Learning framework for CF in this section.
As illustrated in the second example in Figure \ref{fig:illustration} where the domains have mutual user set, we would like to transfer knowledge of those items' records that consistently reflect the user's preferences. Because of our finding that the consistent records have small empirical error and variance, the selection shall consider these two factors.
We embed these two factors into a boosting framework, where the source data with small empirical error and variance receive higher weights since they are consistent with the target data.
This boosting framework models the cross-domain CF from two aspects:
on one hand, we take more care of those mis-predicted target instances;
on the other hand, we automatically identify the consistency of the source domains during the learning process and selective use those source domains with more trustful information.


As shown in Algorithm \ref{algorithm:VPB-TLCF}, in each iteration, we apply base model TGPLSA over weighted instances to build a weak learner $G(\cdot)$ and hypothesis $h^{iter}$. Then to update the source and target item weights, domain level fitness weight $\beta^{s_k}$ is chosen for each source domain $s_k$ based on domain level consistency ~\cite{/ecml/Eaton2008Modeling}. And $\alpha^{iter}$ for base model is also updated, considering empirical errors and variances. Accordingly, the weights of mis-predicted target items are increased and the weights of those less helpful source domains are decreased in each iteration. The final ensemble is given by an additive model, which gives larger weights to the hypotheses with lower errors.
We provide a detailed derivation of STLCF in the rest of this section.

In previous works in collaborative filtering, the mean absolute error (MAE) is usually chosen as the loss function. If we tolerate some prediction error $\tau$, we define:
\begin{equation}
l_1(\X_{* i},\hat \X_{* i})=\left\{ \begin{aligned}
         &-1, & \!\!\!\sum_{x_{ui}\in\X_{* i}}\!|\hat x_{ui}\!-\!x_{ui}| \!\leq\! \tau \cdot nnz(\X_{* i})\\
         &+1, & \!\!\!\sum_{x_{ui}\in\X_{* i}}\!|\hat x_{ui}\!-\!x_{ui}| \!>\! \tau \cdot nnz(\X_{* i})\\
                          \end{aligned} \right.
\end{equation}
where $nnz(\cdot)$ is the number of observed ratings. $\X_{* i}$ and $\hat \X_{* i}$ denote the true values and predictions respectively.
We may also define the item level MAE error for target domain with respect to $\tau$ as:
\begin{eqnarray}\label{eq:mae_loss}
    \epsilon^{d}_i = l_1(\X^{d}_{* i}, \hat \X^{d}_{* i})
\end{eqnarray}
To facilitate the optimization, we consider the following exponential loss for empirical risk minimization:
\begin{eqnarray}\label{eq:exp_loss}
    l_2(i) = l_2(\X^{d}_{* i}, \hat \X^{d}_{* i}) = e^{\epsilon^{d}_i}
\end{eqnarray}
As stated in previous section, the lower variance of empirical errors can provide more confident consistency estimation, we combine these factors and reformulate the loss function:
\begin{eqnarray}\label{eq:vpb_loss}
    \Loss =
    \sum_{i=1}^{n^d} l_2(i) + \gamma\sqrt{\sum_{i>j}^{n^d} (l_2(i)-l_2(j))^2}
\end{eqnarray}
Above all, the model minimize the above quantity for some scalar $\gamma>0$:

Assume that the function of interest $\mathcal{H}$ for prediction is composed of the hypothesis $h^t$ from each weak learner. The function to be output would consist of the following additive model over the hypothesis from the weak learners:
\begin{eqnarray}
    \hat x^d_{ui} = f(x^d_{ui}) = \sum_{t=1} \alpha^t h^t(x^d_{ui})
\end{eqnarray}
where $\alpha^t \in \mathbb{R}^+$.

Since we are interested in building an additive model, we assume that we already have a function $h(\cdot)$. Subsequently, we derive a greedy algorithm to obtain a weak learner $G^t(\cdot)$ and a positive scalar $\alpha^t$ such that $f(\cdot) = h(\cdot) + \alpha^t G^t(\cdot)$.

In the following derivation, for the convince of presentation, we omit the model index $t$, and use $G$ to represent $G^t$, $\alpha$ to represent $\alpha^t$.

By defining $\gamma_1=(1+(n-1)\gamma)$, $\gamma_2=(2-2\gamma)$, $\alpha$,
 $w_i^d=e^{l_1(h(\X^d_{*i}),\X^d_{*i})}$ and $G_i^d=l_1(G(\X^d_{*i}),\X^d_{*i})$, Eq.(\ref{eq:vpb_loss}) can be equivalently posed as optimizing the following loss with respect to $\alpha$:
\begin{eqnarray}\label{eq:add}
    \\
    \nonumber
    \Loss
    &=&\gamma_1(\sum_{i\in I} (w_i^d)^2 e^{2\alpha}+\sum_{i\in J} (w_i^d)^2 e^{-2\alpha})+\gamma_2\!\!\!\!\!\sum_{i>j:i,j\in I}^{n^d}\!\!\!\!\! w_i^d w_j^d e^{2\alpha}\\ \nonumber
    & &+\gamma_2\!\!\!\!\!\sum_{i>j:i,j\in J}^{n^d}\!\!\!\!\! w_i^d w_j^d e^{-2\alpha}+\gamma_2\!\!\!\!\!\!\!\!\sum_{i>j:i\in I,j\in J or i\in J, j\in I}^{n^d}\!\!\!\!\!\!\!\!\!\!\!\! w_i^d w_j^d
\end{eqnarray}
For brevity, we define the following sets of indices as $I=\{i:G^d_i=+1\}$ and $J=\{i:G^d_i=-1\}$. Here $J$ denotes the set of items whose prediction by $G(\cdot)$ falls into the fault tolerable range, while $I$ denotes the rest set. By making the last transformation in Eq.(\ref{eq:add}) equal to zero, we get:
\begin{eqnarray}\label{eq:alpha}
    \\
    \nonumber
    \alpha =\frac{1}{4}\log\left(\frac{(1-\gamma)(\sum_{i\in I}w_i^d)^2+\gamma n^d\sum_{i\in I}(w_i^d)^2}{(1-\gamma)(\sum_{i\in J}w_i^d)^2+\gamma n^d\sum_{i\in J}(w_i^d)^2}\right)
\end{eqnarray}
If we set $\gamma=0$, then it is reduced to the form of AdaBoost:
\begin{eqnarray}\label{eq:adaboost_alpha}
    \\
    \nonumber
    \alpha =\frac{1}{4}\log\left(\frac{(\sum_{i\in I}w_i^d)^2}{(\sum_{i\in J}w_i^d)^2}\right)=\frac{1}{2}\log\left(\frac{(\sum_{i\in I}w_i^d)}{(\sum_{i\in J}w_i^d)}\right)
\end{eqnarray}
Finally, the updating rule for $w_i^d$ is
\begin{eqnarray}\label{eq:wi}
    w_i^d \leftarrow w_i^d e^{(-\alpha G_i^d)}
\end{eqnarray}
And for the instance weight $w_i^d$ in the source domain, we can also adopt the similar updating rule in Eq.(\ref{eq:wi}).

Other than the instance level selection discussed above, we also want to perform the domain level selection to penalize those domains that are likely to be irrelevant, so that the domains with more relevant instances speak loudly.
Following the idea of task-based boosting ~\cite{/aaai/Eatond11}, we further introduce a re-weighting factor $\beta$ for each source domain to control the knowledge transfer. So we formulate the updating rule for $w_i^s$ to be:
\begin{eqnarray}\label{eq:wi_s}
    w_i^s \leftarrow w_i^s e^{(-\alpha G_i^s-\beta)}
\end{eqnarray}
where $\beta$ can be set greedily in proportion to the performance gain of the single source domain transfer learning:
\begin{eqnarray}\label{eq:beta}
    \beta = \frac{\sum w^d_i (\varepsilon_i - \vec{\varepsilon}_i)}{||\W^d||_1}
\end{eqnarray}
where $\varepsilon_i$ is the training error of the transfer learning model, and $\vec{\varepsilon}_i$ is the training error of the non-transfer learning model, which utilizes only the observed target domain data.

\section{Experiments}


\begin{table}[!tbp]
\caption{\footnotesize Datasets in our experiments.}
\label{tbl:notationDataset}
\begin{footnotesize}
\begin{center}
\begin{tabular}{ c | c | c |c}
\hline\hline
Notation & Data Set & Data Type & Instances No. \\
\hline \hline
D1 & Douban Music & Rating [1,5]& $1.2 \times 10^6$\\
D2 & Douban Book & Rating [1,5]& $5.8 \times 10^5$ \\
D3 & Douban Movie & Rating [1,5]& $1.4 \times 10^6$ \\
D4 & Netflix & Rating [1,5]& $1.8 \times 10^4$\\
D5 & Wikipedia & Editing Log & $1.1 \times 10^6$\\
D6 & IMDB & Hyperlink & $5.0 \times 10^3$\\
\hline\hline
\end{tabular}
\end{center}
\end{footnotesize}
\vspace{-7mm}
\end{table}

\subsection{Data Sets and Experimental Settings} \label{sec:DataSets}
We evaluate the proposed method on four data sources:
Netflix\footnote{http://www.netflix.com},
Douban
IMDB\footnote{http://www.imdb.com},
and Wikipedia\footnote{http://en.wikipedia.org} user editing records.
The Netflix rating data contains more than $100$ million ratings with values in $\{1, 2, 3, 4, 5\}$, which are given by more than $4.8\times 10^5$ users
on around $1.8\times 10^4$ movies.
Douban contains movie, book and music recommendations, with rating values also in $\{1, 2, 3, 4, 5\}$.
IMDB hyperlink graph is employed as a measure of similarity between movies.
In the graph, each movie builds links to its $10$ most similar movies.
The Wikipedia user editing records provide a $\{0, 1\}$ indicator of whether a user concerns or not about a certain movie.

The data sets used in the experiments are described as follows. For Netflix, to retain the original features of the users while keeping the size of the data set suitable for the experiments, we sampled a subset of $10,000$ users.
In Douban data sets, we obtained $1.2 \times 10^6$ ratings on $7.5 \times 10^3$ music, $ 5.8 \times 10^5 $ ratings on $3.5 \times 10^3 $ books, and $1.4 \times 10^6 $ ratings on $8 \times 10^3$ movies, given by $1.1 \times 10^4 $ users.
For both the IMDB data set and the Wikipedia data set, we filtered them by matching the movie titles in both the Netflix and the Douban Movie data sets. After pre-processing, the IMDB hyperlink data set contains $\sim 5 \times 10^3$ movies. The Wikipedia user editing records data set has $1.1 \times 10^6$ editing logs by $8.5 \times 10^3$ users on the same $\sim 5 \times 10^3$ movies as IMDB data set.
To present our experiments, we use the shorthand notations listed in Table \ref{tbl:notationDataset} to denote the data sets.

We evaluate the proposed algorithm on five cross-domain recommendation tasks, as follows:
\begin{itemize}[noitemsep,topsep=0pt,parsep=0pt,partopsep=0pt]
\item The first task is to simulate the cross-domain collaborative filtering, using the Netflix data set. The sampled data is partitioned into two parts with disjoint sets of movies but identical set of users. One part consists of ratings given by $8,000$ movies with $1.6\%$ density, which serves as the source domain. The remaining $7,000$ movies are used as the target domain with different levels of sparsity density.
\item The second task is a real-world cross-domain recommendation, where the source domain is Douban Book and the target domain is Douban Movie. In this setting, the source and the target domains share the same user set but have different item sets.
\item The third task is constructed with Netflix and Douban data. There are about $6,000$ shared movies in Netflix and Douban Movie. We extract the ratings on shared movies from Netflix and Douban Movie. Then we get $4.9 \times 10^5$ ratings from Douban given by $1.2 \times 10^4$ users with density $0.7\%$, and $10^6$ ratings from Netflix given by $10^4$ users with density $1.7\%$. The goal is to transfer knowledge from the Netflix data set to Douban Movie. In this task, item set is identical across domains but user sets are totally different.
\item The fourth task is to evaluate the effectiveness of the proposed algorithm under the context of multiple source domains. It uses both Douban Music and Douban Book as the source domains and transfer knowledge to Douban Movie domain.
\item The fifth task varies the type of source domains. It utilizes the Wikipedia user editing records and IMDB hyperlink graph, together with Douban Movie as the source domains to perform rating predictions on the Netflix movie data set.
\end{itemize}

For evaluation, we calculate the Root Mean Square Error (RMSE) on the heldout $\sim 30\%$ of the target data:
\begin{eqnarray}
    RMSE &=& \sqrt{ \sum_{ (u, i, x_{ui}) \in T_E } (x_{ui} - \hat{x}_{ui})^2 / |T_E|} \nonumber
\end{eqnarray}
where $x_{ui}$ and $\hat{x}_{ui}$ are the true and predicted
ratings, respectively, and $|T_E|$ is the number of test ratings.

\begin{table*}[!tbp]
\caption{\footnotesize Prediction performance of STLCF and the baselines.}
\label{tbl:TLCF}
\begin{footnotesize}
\begin{center}
\begin{tabular}{ c || c || c || c c || c c || c  c } \hline\hline

 \multirow{2}{*}{Datasets}
& \multirow{1}{*}{Source}
& \multirow{1}{*}{Target}
& \multicolumn{2}{c ||}{Non-TL}
& \multicolumn{2}{c ||}{Non-Selective TL}
& \multicolumn{2}{c }{\bf Selective TL}\\
\cline{4-9}
&sparseness
&sparseness
& \multicolumn{1}{c }{GPLSA}
& \multicolumn{1}{c||}{PMF}
& \multicolumn{1}{c }{TGPLSA}
& \multicolumn{1}{c||}{CMF}
& \multicolumn{1}{c }{STLCF(E)}
& \multicolumn{1}{c}{STLCF(EV)}\\
\hline \hline
D4(Simulated) & \multirow{3}{*}{1.6\%} & 0.1\% & 1.0012 & 0.9993 & 0.9652 & 0.9688 & 0.9596 & \textbf{0.9533} \\
\cline{3-9}
to      &                        & 0.2\% & 0.9839 & 0.9814 & 0.9528 & 0.9532 & 0.9468 & \textbf{0.9347} \\
\cline{3-9}
D4(Simulated) &                        & 0.3\% & 0.9769 & 0.9728 & 0.9475 & 0.9464 & 0.9306 & \textbf{0.9213} \\
\hline \hline
& \multirow{3}{*}{1.5\%} & 0.1\% & 0.8939 & 0.8856 & 0.8098 & 0.8329 & 0.7711 & \textbf{0.7568} \\
\cline{3-9}
D2 to D3&                        & 0.2\% & 0.8370 & 0.8323 & 0.7462 & 0.7853 & 0.7353 & \textbf{0.7150} \\
\cline{3-9}
&                        & 0.3\% & 0.7314 & 0.7267 & 0.7004 & 0.7179 & 0.6978 & \textbf{0.6859} \\
\hline \hline
& \multirow{3}{*}{1.7\%} & 0.1\% & 0.8939 & 0.8856 & 0.8145 & 0.8297 & 0.7623 & \textbf{0.7549} \\
\cline{3-9}
D4 to D3 &                        & 0.2\% & 0.8370 & 0.8323 & 0.7519 & 0.7588 & 0.7307 & \textbf{0.7193} \\
\cline{3-9}
&                        & 0.3\% & 0.7314 & 0.7267 & 0.7127 & 0.7259 & 0.6982 & \textbf{0.6870} \\
\hline \hline

\end{tabular}
\end{center}
\end{footnotesize}
\vspace{-7mm}
\end{table*}

\subsection{STLCF and Baselines Methods}
We implement two variations of our STLCF method.
STLCF(E) is an STLCF method that only take training error into consideration when performing selective transfer learning.
STLCF(EV) not only considers training error, but also utilizes the empirical error variance.
To demonstrate the significance of our STLCF, we selected the following baselines\footnote{Parameters for these baseline models are fine-tuned via cross validation.}:
{\bf PMF}~\cite{/nips/SalakhutdinovM07} is a recently proposed method for missing value prediction. Previous work showed that this method worked well on the large, sparse and imbalanced data set.
{\bf GPLSA}~\cite{DBLP:conf/sigir/Hofmann03} is a classical non-transfer recommendation algorithm.
{\bf CMF}~\cite{/kdd/SinghG08} is proposed for jointly factorizing two matrices. Being adopted as a transfer learning technique in several recent works, CMF has been proven to be an effective cross-domain recommendation approach.
{\bf TGPLSA} is an uniformly weighted transfer learning model, which utilize all source data to help build the target domain model. It is used as one of the baselines because we adopt it as the base model of our boosting-based selective transfer learning framework.

\subsection{Experimental Results}

\subsubsection{Performance Comparisons}
We test the performance of our STLCF methods against the baselines.
The results of the collaborative filtering tasks under three different target domain sparseness are shown in Table \ref{tbl:TLCF}.

First, we observe that the non-transfer methods, i.e. GPLSA and PMF, fail to give accurate predictions, especially when the target domain is severely sparse.
With the help of source domains, the (non-selective) transfer learning methods with equally weights on the source domains, like TGPLSA and CMF, can increase the accuracy of the rating predictions. And our selective transfer learning methods (i.e., STLCF(E) and STLCF(EV)) can do even better.
The fact that our STLCF outperforms others is expected because by performing the {\em selective} knowledge transfer, we use the truly helpful source domain(s), which is designed to handle the sparseness issue in CF problems.

Second, comparing the two non-selective TLCF methods with the other two selective TLCF methods, we observe that on the last two real world tasks (D2 to D3 and D4 to D3) when the target domain is extremely sparse (say 0.1\%), the improvement of accuracy achieved by our STLCF methods against the non-selective transfer learning methods is much more significant than it does on the simulation data set based on Netflix (D4 to D4).
Notice that the inconsistency of the target domain and the source domains on the simulation data sets is much smaller than that on the real-world cases. The experiment results show that our STLCF algorithm is effective in handling the inconsistency between the sparse target domain and the source domains.

Third, we notice that some factors, like empirical error variance, may affect the prediction. In Table \ref{tbl:TLCF}, we compare our two STLCF methods, i.e., STLCF(E) and STLCF(EV) when the target domain sparsity is $0.1\%$. We can find that on the task ``D2 to D3", i.e., Douban Book to Movie, STLCF(EV) is much better than STLCF(E). But on the task ``D4(Simulated) to D4(Simulated)", the improvement of STLCF(EV) is not so significant against STLCF(E). These observations may be due to the domain consistency.
For the tasks ``D4(Simulated) to D4(Simulated)", both the source and target entities are movie ratings from Netflix data set, while the task ``D2 to D3" tries to transfer the knowledge from a book recommendation system to the moive recommendation system, which may contain some domain specific items.
When the target domain is very sparse, i.e. the user's ratings on the items are rare, there are chances to get high prediction accuracy occasionally on the observed data with a bad model on the source domains that are inconsistent with target domain. In this case, it is important to consider the variance of empirical error as well. Comparing to STLCF(E), STLCF(EV), which punishes the large variance, can better handle the domain inconsistency in transfer learning, especially when the target domain is sparse.

\subsubsection{Results on Long-Tail Users}
\begin{table} [t]
\begin{scriptsize}
\caption{\footnotesize Prediction performance of STLCF for Long-Tail Users on the D2 to D3 task. STLCF(E) does not punish the large variance of empirical error, while STLCF(EV) does.}
\label{tbl:tail}
\begin{center}
\begin{tabular}{l||c|| cc||cc}
\hline\hline
\small{Ratings} & \multirow{1}{*}{\multirow{2}{*}{\!\small{Non-TL}}} & \multicolumn{2}{c ||}{\multirow{2}{*}{\!\small{Non-Selective TL}}} & \multicolumn{2}{c}{\!\small{\bf Selective TL}}\\
\small{per} & & & & \multicolumn{2}{c}{\small {\bf i.e. STLCF}}\\
\cline{2-6}
\small{user} & \small{GPLSA} & \small{TGPLSA} & \small{CMF} & \small{(E)} & \small{(EV)}\\
\hline\hline
1-5   & 1.1942 & 0.9294 & 0.9312 & 0.8307 & \textbf{0.8216}\\\hline
6-10  & 0.9300 & 0.7859 & 0.7929 & 0.7454 & \textbf{0.7428}\\\hline
11-15 & 0.8296 & 0.7331 & 0.7390 & \textbf{0.7143} & 0.7150\\\hline
16-20 & 0.7841 & 0.7079 & 0.7113 & \textbf{0.7042} & 0.7050\\\hline
21-25 & 0.7618 & 0.6941 & 0.6947 & 0.6942 & \textbf{0.6910}\\\hline
26-30 & 0.7494 & 0.6918 & 0.6884 & 0.6917 & \textbf{0.6852}\\\hline
31-35 & 0.7370 & 0.6909 & 0.6911 & 0.6915 & \textbf{0.6818}\\\hline
36-40 & 0.7281 & 0.6896 & 0.6856 & 0.6907 & \textbf{0.6776}\\\hline
41-45 & 0.7219 & 0.6878 & 0.6821 & 0.6890 & \textbf{0.6740}\\\hline
46-50 & 0.7187 & 0.6881 & 0.6878 & 0.6800 & \textbf{0.6734}\\
\hline
\hline
\end{tabular}
\end{center}
\end{scriptsize}
\vspace{-7mm}
\end{table}
To better understand the impact of STLCF with the help of the source domain, we conduct a fine-grained analysis on the performance improvement on Douban data sets, with Douban Book as source domain and Douban Movie as target domain.
The results on different user groups in the target domain are shown in Table \ref{tbl:tail}.
First, we observe that the STLCF models, i.e., STLCF(E) and STLCF(EV) can achieve significantly better results on those long-tail users who have very few ratings in historical logs.
Such fact implies that our STLCF methods could handle the long-tail users that really need a fine-grained analysis when performing knowledge transfer from source domains.
Current CF models without any fine-grained analysis on the specific users usually fail to capture the preferences of the long-tail users, while our STLCF methods work well because they can selectively augment the weight of the corresponding source domain instances with respect to those long-tail cases at both instance level and domain level.
Second, STLCF(EV) works better than STLCF(E) on those non-long-tail users, i.e., with more than 25 ratings per user in the historical log. This is expected because users with more ratings can benefit more from the error variance analysis to avoid negative knowledge transfer.
\begin{table*}[t]
\caption{\footnotesize Prediction performance of STLCF with multiple source domains containing much irrelevant information.}
\begin{footnotesize}
\label{tbl:mdomainshyb}
\begin{center}
\begin{tabular}{ c | c || c || c | c | c | c | c }
\hline\hline
\multicolumn{2}{c ||} {Source Domain: }  & None  & D3 & D3 \& D5 & D3 \& D6 & D5 \& D6 & D3 \& D5 \& D6\\
\hline\hline
 \multirow{1}{*}{Target} & 0.1\% & 0.9983 & $0.9789$& $0.9747$& $0.9712$& $0.9923$&$\textbf{0.9663}$\\
\cline{2-8}
 \multirow{1}{*}{(D4)} & 0.2\% & 0.9812 & $0.9625$& $0.9583$&$0.9572$ & $0.9695$ & $\textbf{0.9505}$\\
\cline{2-8}
 \multirow{1}{*}{sparseness} & 0.3\%& 0.9703 & $0.9511$& $0.9409$& $0.9464$& $0.9599$ &$\textbf{0.9383}$\\
\hline\hline
\end{tabular}
\end{center}
\end{footnotesize}
\vspace{-7mm}
\end{table*}

\begin{table}[!tbp]
\caption{\footnotesize Prediction performance of STLCF with multiple source domains (Douban).}
\label{tbl:mdomainsDB}
\begin{footnotesize}
\begin{center}
\begin{tabular}{ c |c || c || c | c | c }
\hline\hline
\multicolumn{2}{c ||} {Source Domain:} & None & D1 & D2 & D1 \& D2\\
\hline\hline
\multirow{1}{*}{Target} & 0.1\% & 0.8856& $0.7521$ & $0.7568$ & $\textbf{0.7304}$\\
\cline{2-6}
\multirow{1}{*}{(D3)} & 0.2\% & 0.8323 & $0.7163$ & $0.7150$ & $\textbf{0.6904}$\\
\cline{2-6}
\multirow{1}{*}{sparseness} & 0.3\%& 0.7267 & $0.6870$ & $0.6859$ & $\textbf{0.6739}$\\
\hline\hline
\end{tabular}
\end{center}
\end{footnotesize}
\vspace{-10mm}
\end{table}

\subsubsection{STLCF with Multiple Source Domains}
We apply STLCF(EV) on the extremely sparse target movie domain, with two sets of source domains: one is composed of Douban Music and Douban Book, the other is composed of Douban Movie, IMDB hyperlink graph and Wikipedia user editing records. The results are in Table \ref{tbl:mdomainsDB} and Table \ref{tbl:mdomainshyb} respectively. We demonstrate our STLCF method can utilize multiple source domains of various types by handling the inconsistency between the target and the source domains.

First, for the Douban experiments shown in Table \ref{tbl:mdomainsDB},
we observe that comparing to only using either Douban Book or Douban Music as source domain, there are significant improvements when both of them are used.
The result is expected because each of the source domains has its own parts of effective information for the target domain.
For example, a user who show much interests in the movie ``The Lord of the Rings'' may have consistent preferences in its novel. In this case, with the help of more auxiliary sources, better results are expected.

Second, we explore the generalization of the choices of source domains by introducing domains like Wikipedia user editing records and IMDB hyperlink graph, which are not directly related to the target domain but still contain some useful information in helping the target task (Netflix rating prediction). The results are shown in Table~\ref{tbl:mdomainshyb}.
Comparing the results of the experiment that uses no source domain (non-transfer) to those that use source domains D5 \& D6, we observe that although the Wikipedia user editing records or IMDB hyperlink graph is not closely related to the target domain and can hardly be adopted as source domains by previous transfer learning techniques, our STLCF method can still transfer useful knowledge successfully.
In addition, comparing the results of the experiment that uses single source domain D3 to those that use source domains D3 \& D5, D3 \& D6, or D3 \& D5 \& D6, we find that the Wikipedia user editing records or IMDB hyperlink graph could provide some useful knowledge that is not covered by the related movie source domains. Despite of the noise and heterogeneous setting, our STLCF method can still utilize these source domains to help the target domain tasks. As we have discussed in Section \ref{sec:STLCF}, our STLCF performs selective transfer learning at both domain level and instance level.
On one hand, the domain level selective transfer can block the noisy information globally. As we can see, D5 \& D6 are noisy and therefore contain much data that are inconsistent with the observed data in the target domain, therefore the overall transfer of D5 \& D6 is penalized.
On the other hand, the instance level selective transfer learning can further eliminate the affections of those irrelevant source instances.

Above all, our STLCF is highly adaptive to utilize source domains that are relatively inconsistent with the target domain, even when the target domain is rather sparse.

\subsubsection{Parameters Analysis of STLCF}
There are two parameters in our STLCF, i.e., the prediction error threshold $\tau$ and the empirical error variance weight $\gamma$.
Since $\tau$ and $\gamma$ are independent, we fix one and adjust another.

\begin{figure}[t]
\begin{minipage}[t]{0.45\linewidth}
\includegraphics[width=1.8in]{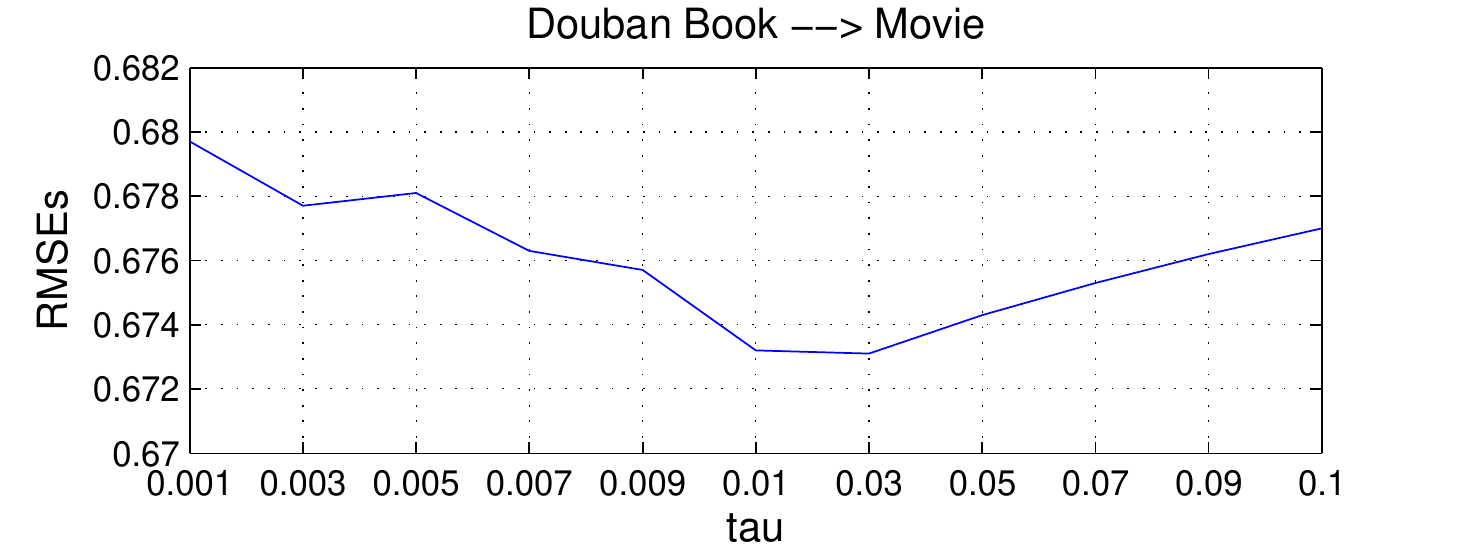}
\end{minipage}
\hspace{0.1in}
\begin{minipage}[t]{0.45\linewidth}
\includegraphics[width=1.8in]{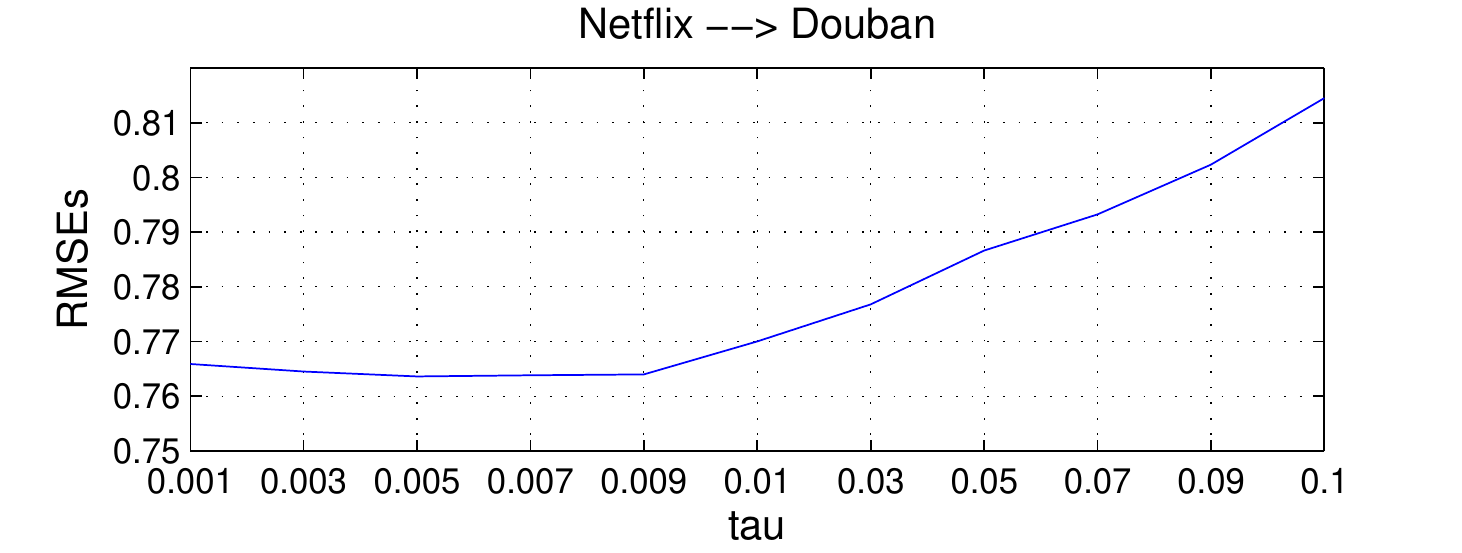}
\end{minipage}
\vspace{-3mm}
\caption{\footnotesize Change of the RMSEs with different $\tau$s.}\label{fig:tau}
\end{figure}

\begin{figure}[!t]
\begin{minipage}[t]{0.45\linewidth}
\includegraphics[width=1.8in]{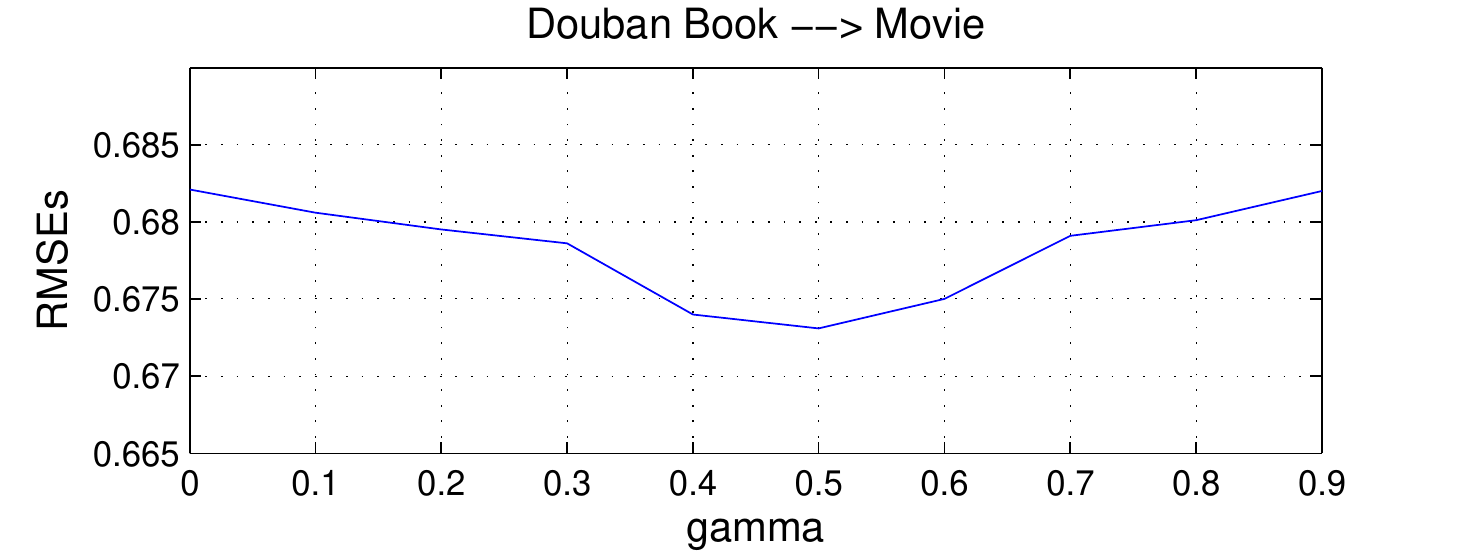}
\end{minipage}
\hspace{0.1in}
\begin{minipage}[t]{0.45\linewidth}
\includegraphics[width=1.8in]{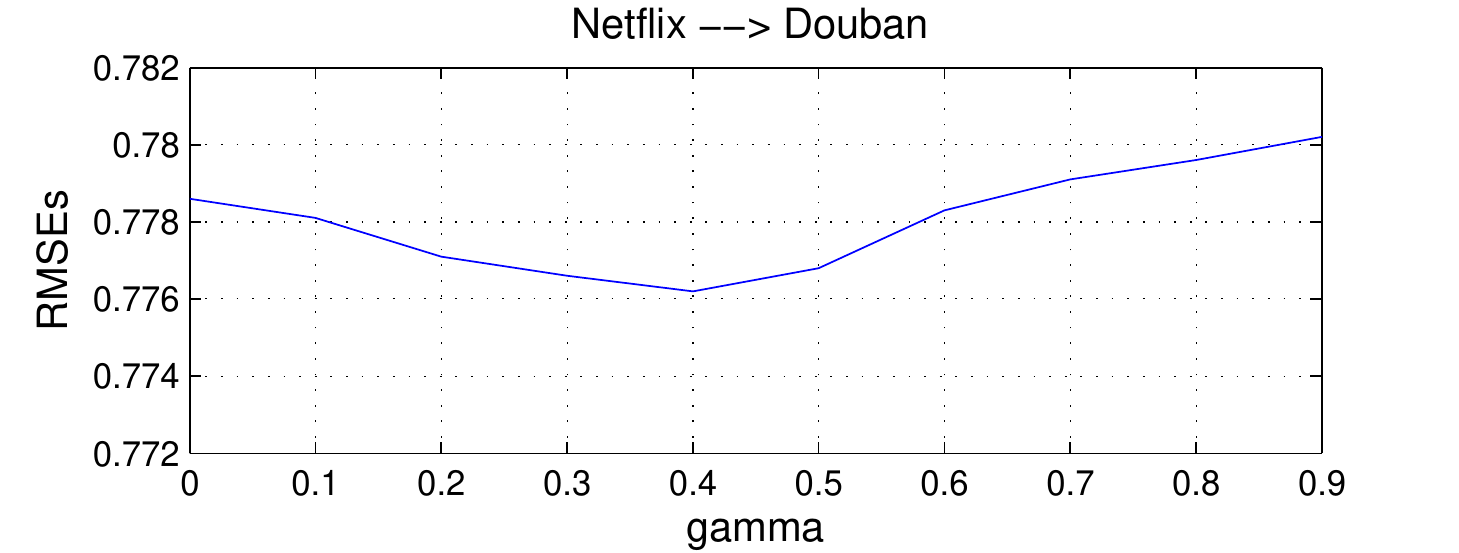}
\end{minipage}
\vspace{-3mm}
\caption{\footnotesize Change of the RMSEs with different $\gamma$s.}\label{fig:gamma}
\vspace{-5mm}
\end{figure}

We fix the empirical error variance weight to be $\gamma=0.5$ and adjust the parameter $\tau$. Based on our results shown in Figure \ref{fig:tau}, the model has good performance when $\tau$ is of order $10^{-2}$.
We also tuned the parameter $\gamma$, which balances the empirical error and its variance. We fix the prediction error threshold to be $\tau=0.03$ in tuning $\gamma$. As shown in Figure \ref{fig:gamma}, when we vary the parameter $\gamma$ from $0$ to $1$, the best choices of $\gamma$ are found to be around $0.4-0.5$.

\subsubsection{Convergence and Overfitting Test}
\begin{figure}[t]
\begin{minipage}[t]{0.45\linewidth}
\includegraphics[width=1.7in]{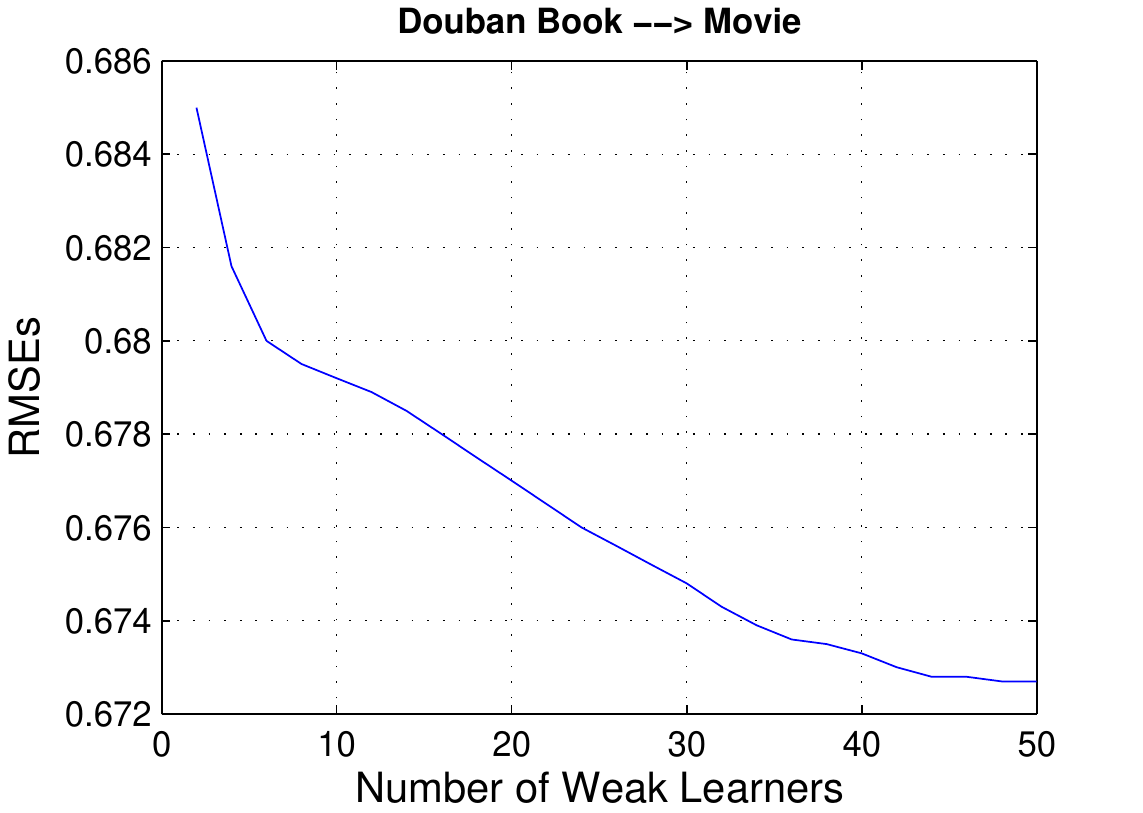}
\end{minipage}
\hspace{0.1in}
\begin{minipage}[t]{0.45\linewidth}
\includegraphics[width=1.7in]{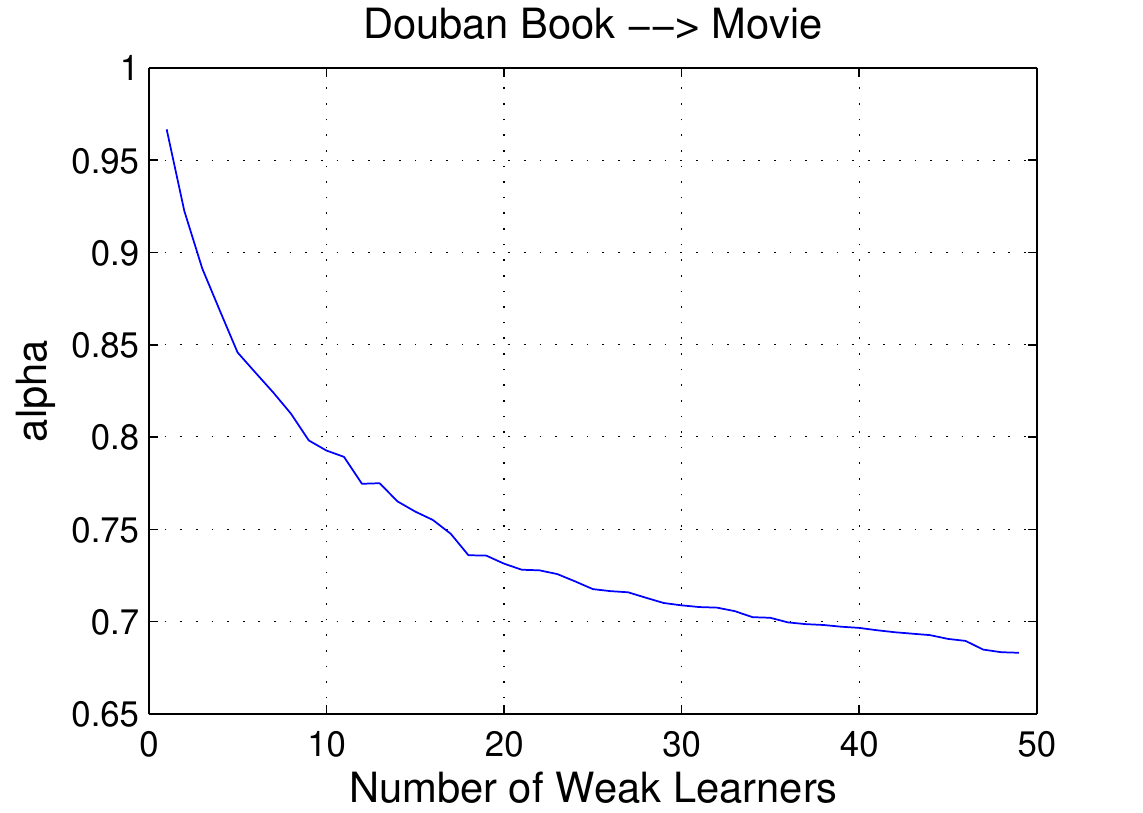}
\end{minipage}
\caption{\footnotesize Change of the RMSEs and $\alpha$s when more and more weak learners join in the committee.}\label{fig:converge}
\vspace{-5mm}
\end{figure}

Figure \ref{fig:converge} shows the RMSEs of STLCF(EV) as the number of weak learners changes on the Douban Book to Movie task. From the figure on the left, we observe that STLCF(EV) converges well after 40 iterations. We can also find that the corresponding $\alpha$ also converge to around 0.68 after 40 iterations as well.

\begin{figure}
\begin{minipage}[t]{0.95\linewidth}
\includegraphics[width=3.3in]{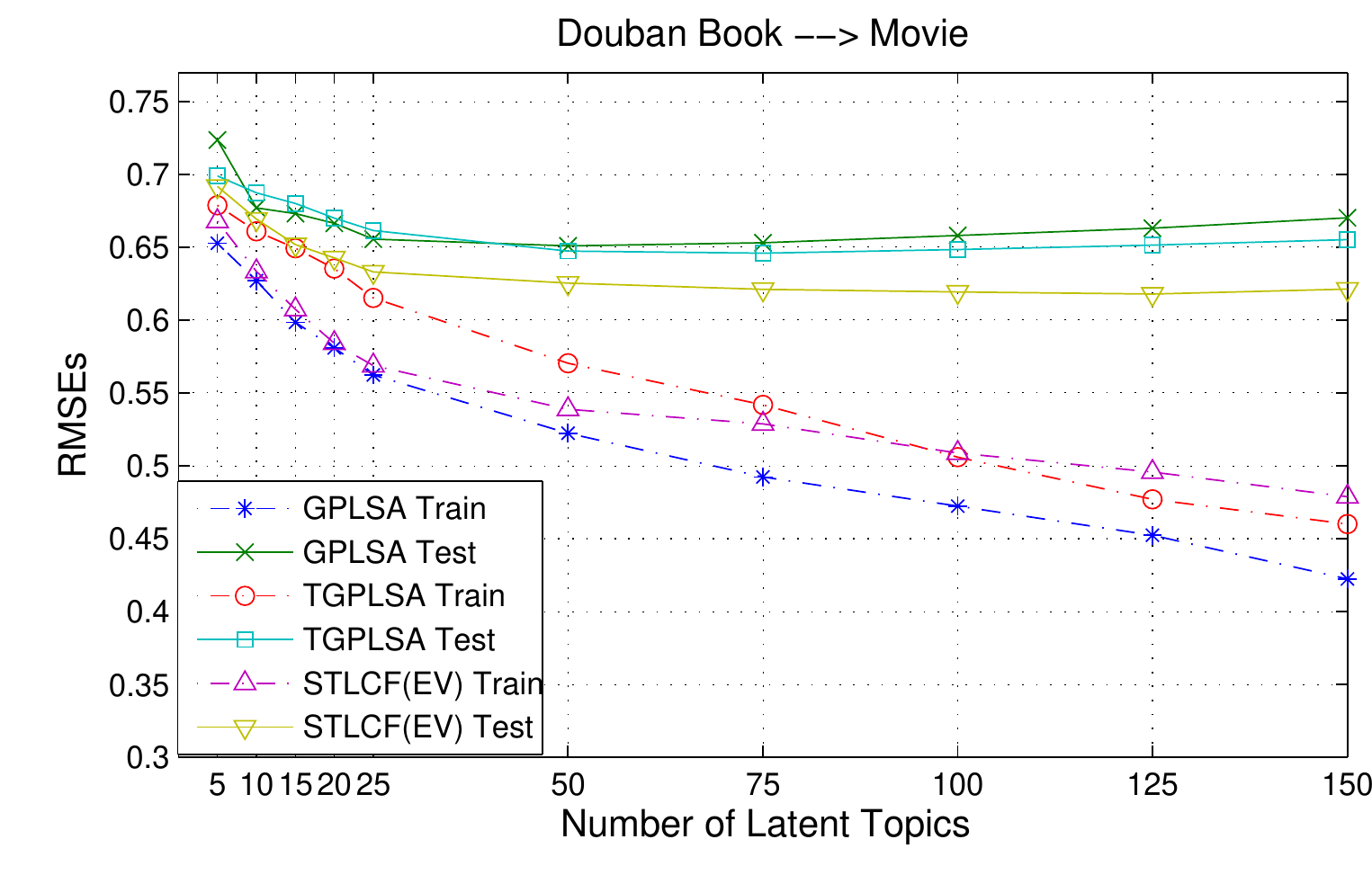}
\end{minipage}
\vspace{-3mm}
\caption{\footnotesize Change of the RMSEs with different numbers of latent topics.}\label{fig:k}
\vspace{-5mm}
\end{figure}

The number of latent topics of the base learner TGPLSA reflects the model's ability to fit training data. When we keep increasing the number of latent topics, the model tends to better fit the training data. But if the number of latent topics is too large, the model may suffer from overfitting.
We investigate the overfitting issue by plotting the training and testing RMSEs of the non-transfer learning model GPLSA, the non-selective transfer learning model TGPLSA and our selective transfer learning model STLCF(EV) over different numbers of latent topics in Figure \ref{fig:k}. The data sparsity for the target domain is around 0.3\%.

We can observe that comparing to our STLCF, the training RMSEs of GPLSA and TGPLSA decrease faster, while their testing RMSEs go down slower. When $k$ is about $50$, the testing RMSEs of GPLSA start to go up. And for TGPLSA, its testing RMSEs also go up slightly when $k$ is larger than $75$. But the testing RMSEs of our STLCF keep decreasing until $k=125$ and even when $k$ is larger than $125$, the raise of our STLCF's testing RMSEs is not obvious. Clearly when the target domain is very sparse, our STLCF method is more robust against the overfitting, by inheriting the advantage from boosting techniques and the fine-grained selection on knowledge transfer.

\section{Related Works}

\begin{table}[!tbp]
\caption{\footnotesize Overview of STLCF in a big picture of collaborative filtering.}
\begin{footnotesize}
\label{tbl:relatedW}
\begin{center}
\begin{tabular}{ c || c | c }
\hline\hline
& Selective & Non-Selective \\
\hline\hline
\multirow{1}{*} {Transfer} & \textbf{\bf \em STLCF} & RMGM~\cite{/ijcai/libin09}, CMF~\cite{/kdd/SinghG08},\\
\multirow{1}{*} {Learning} & & TIF~\cite{/aaai/WPan12}, etc.\\
\cline{1-3}
\multirow{1}{*} {Non-Transfer} & -- & MMMF~\cite{/icml/RennieS05}, GPLSA~\cite{DBLP:conf/sigir/Hofmann03}, \\
\multirow{1}{*} {Learning} &  & PMF~\cite{/nips/SalakhutdinovM07}, etc.\\
\hline\hline
\end{tabular}
\end{center}
\end{footnotesize}
\vspace{-5mm}
\end{table}

The proposed {\textbf S}elective {\textbf T}ransfer {\textbf L}earning for {\textbf C}ollaborative {\textbf F}iltering (STLCF) algorithm is most related to the works in collaborative filtering.
In Table~\ref{tbl:relatedW}, we summarize the related works under the collaborative filtering context.
To the best of our knowledge, no previous work for transfer learning on collaborative filtering has ever focused on the fine-grained analysis of consistency between source domains and the target domain, i.e., the selective transfer learning.

{\bf Collaborative Filtering} as an intelligent component in recommender systems has gained extensive interest in both academia and industry.
Various models have been proposed, including factorization models~\cite{/computer/yehuda09matrix, /aaai/WPan12,paterek07,/tist/LibFM-TIST12, /icml/RennieS05},
probabilistic mixture models~\cite{hofmann04cf,jin:decoupled},
Bayesian networks~\cite{pennock00pd} and restricted Boltzman machines~\cite{/icml/SalakhutdinovMH07}.
However, most of the previous work would suffer from overfitting to the small set of observed data. In this paper, we introduce the concept of selective transfer learning to better tackle the overfitting and data sparseness issue.

{\bf Transfer Learning}~\cite{/tkde/sinno09survey} utilizes data sets from related but different domains to build model for a target application domain. A few works on transfer learning are in the context of collaborative filtering.
Mehta and Hofmann~\cite{/ki/bhaskar06cross} consider the scenario involving two systems with shared users and use manifold alignment methods to jointly build neighborhood models for the two systems. They focus on making use of an auxiliary recommender system when only part of the users are aligned, which does not distinguish the consistency of users' preferences among the aligned users.
Li \etal~\cite{/icml/libin09} designed a regularization framework to transfer knowledge of cluster-level rating patterns, which does not make use of the correspondence between source and target domains.

Recently, researchers propose the MultiSourceTrAdaBoost~\cite{/cvpr/YaoD10} to allow automatically selecting the appropriate data for knowledge transfer from multiple sources. The newest work TransferBoost~\cite{/aaai/Eatond11} was proposed to iteratively construct an ensemble of classifiers via re-weighting source and target instance via both individual and task-based boosting. Moreover, EBBoost~\cite{/jmlr/ShivaswamyJ10a} suggests weight the instance based on the empirical error as well as its variance. However so far, the works limit to the classification tasks. Our work is the first to systematically study {\em selective} knowledge transfer in the settings of collaborative filtering. Besides, we propose the novel factor - variance empirical error that is proven to be of much help in solving the real world CF problems.

\section{Conclusions}

In this paper, we proposed to perform {\em selective} knowledge transfer for CF problems and came up with a systematical study on how the factors such as variance of empirical error could leverage the selection.
We found although empirical error is effective to model the consistency across domains, it would suffer from the sparseness problem in CF settings. By introducing a novel factor - variance of empirical error to measure how trustful this consistency is, the proposed criterion can better identify the useful source domains and the helpful proportions of each source domain.
We embedded this criterion into a boosting framework to transfer the most useful information from the source
domains to the target domain.
The experimental results on real-world data sets showed that our selective transfer learning solution performs significantly better than several state-of-the-art methods at various sparsity levels.
Furthermore, comparing to existing methods, our solution works well on long-tail users and is more robust to overfitting.

\bibliographystyle{abbrv}
\bibliography{luz,TLcf,evan}

\end{document}


\title{Selective Transfer Learning for Cross Domain Recommendation \\ - Appendix}

\date{}

\maketitle
\appendix

\section{Expectation - Maximization algorithm for TGPLSA}
As we discussed in the paper, we obtain the objective function of TGPLSA as:
\begin{eqnarray}\label{eq:tplsa_likelihood}
\\
\nonumber
\Loss = - \sum_{r \in R} ( \lambda \sum_{x_{ui}^s \in \X^s} \log ( w_i^s \cdot Pr(x_{ui}^s = r \mid u^s,i^s ; \theta^s) )  \\\nonumber
+ (1-\lambda) \sum_{x_{ui}^d \in \X^d} \log ( w_i^d  \cdot Pr(x_{ui}^d = r \mid u^d,i^d ; \theta^{d}) ) )
\end{eqnarray}

We adopt the expectation-maximization (EM) algorithm, a standard method for statistical inference, to find the maximum log-likelihood estimates of Eq. (\ref{eq:tplsa_likelihood}).
For the convenience of presentation, in the following derivation, we denote $x_{ui}^l\!=\!r, u^l, i^l$ as $x_{ui}^l$.

In the E-step, we compute the posterior probabilities of the hidden variables, as follows:
\begin{eqnarray}\label{eq:p_z_r}
\\
\nonumber
Pr(z \mid x_{ui}^s) \!=\! \frac{Pr(x_{ui}^s \mid z) Pr(z \mid u^s)}{\sum_{z^{'}} Pr(x_{ui}^s \mid z^{'}) Pr(z^{'} \mid u^s)}\\
\nonumber
Pr(z \mid x_{ui}^d) \!=\! \frac{Pr(x_{ui}^d \mid z) Pr(z \mid u^d)}{\sum_{z^{'}} Pr(x_{ui}^d \mid z^{'}) Pr(z^{'} \mid u^d)}
\end{eqnarray}

Following the similar idea of TPLSA~\cite{/sigir/XueDYY08}, we assume that a particular user's preferences on the latent topics are the same across different domains. Also notice that $Pr(z \mid u^d)$ and $Pr(z \mid u^s)$ reflect the users' preferences on the latent topics. Thus, we set $Pr(z \mid u^d)$ to be the same as $Pr(z \mid u^s)$ to build an information bridge for knowledge transfer across different systems.

In the M-step, we solve the maximization problem via the following equation:
\begin{eqnarray}\label{eq:p_z_u}
\\
\nonumber
Pr(z|u)\!=\!\frac{\sum_{l \in \{s,d\}}\!\!\lambda_l\! \!\sum_{r \in R}\!\!\sum_{x_{ui}^l \in \X^l_{u*}}\!\!w^l_i Pr(z | x_{ui}^l)  }{\sum_{z^{'}} (\sum _{l \in \{s,d\}}\!\!\lambda_l\!\!\sum_{r \in R}\!\!\sum_{x_{ui}^l \in \X^l_{u*}}\!\! w^l_i Pr(z^{'} | x_{ui}^l) ) }
\end{eqnarray}
And we obtain the following equations for the parameters of the Gaussian distribution:
\begin{eqnarray}\label{eq:p_mu}
\mu_{iz}^l \!=\! \frac{\sum_r\sum_{x_{ui}^l \in \X^l_{*i}} r \cdot Pr(z \mid x_{ui}^l)}{\sum_r\sum_{x_{ui}^l \in \X^l_{*i}} Pr(z \mid x_{ui}^l)}
\end{eqnarray}
\begin{eqnarray}\label{eq:p_sigma}
\\
\nonumber
(\sigma_{iz}^{l})^2 \!=\! \frac{\sum_r\sum_{x_{ui}^l \in \X^l_{*i}} (r-\mu_{iz})^2 Pr(z \mid x_{ui}^l)}{\sum_r\sum_{x_{ui}^l \in \X^l_{*i}} Pr(z \mid x_{ui}^l)}
\end{eqnarray}
After proceeding the EM algorithm by alternating E-step using Eq. (\ref{eq:p_z_r}) with M-step using Eq. (\ref{eq:p_z_u}), (\ref{eq:p_mu}) and (\ref{eq:p_sigma}), finally we compute the expected result in the target domain as
\begin{eqnarray}\label{eq:pred}
\\
\nonumber
\hat x_{ui}^d\!=\!E[r \mid u,i] \!=\! \int_R r Pr(r \mid u, i) dr \!\approx\! \sum_{z} Pr(z \mid u) \mu_{i,z}^d
\end{eqnarray}

\bibliographystyle{abbrv}
\bibliography{luz,TLcf,evan}